\documentclass[sigconf, nonacm, balance=false]{acmart}
\AtBeginDocument{%
  }

\usepackage{tikz}
\usepackage{float}
\usetikzlibrary{positioning, arrows.meta}
\usepackage{adjustbox}
\usepackage{booktabs}
\usepackage{pdflscape}
\usepackage{makecell}

\begin{document}

\settopmatter{printfolios=true}
\title{When Do Data-Driven Systems Exhibit the Capability to Infer?}

\author{Maximilian Poretschkin} 
\affiliation{
  \institution{Fraunhofer Institute for Intelligent Analysis and Information Systems (IAIS)}
  \institution{University of Bonn}
  \institution{Lamarr Institute for Machine Learning and Artificial Intelligence}
  \city{Sankt Augustin, Bonn}
  \state{}
  \country{Germany}
}

\email{maximilian.poretschkin@iais.fraunhofer.de}
\author{Tabea Naeven} 

\affiliation{%
  \institution{Fraunhofer Institute for Intelligent Analysis and Information Systems (IAIS)}
  \city{Sankt Augustin}
  \state{}
  \country{Germany}
}
\email{tabea.naeven@iais.fraunhofer.de}

\renewcommand{\shortauthors}{Poretschkin and Naeven}

\begin{abstract}
The European AI Act is the first comprehensive regulation of artificial intelligence (AI), setting out extensive obligations, particularly for so-called high-risk and general-purpose AI systems. A key distinguishing feature of AI systems under the AI Act is the capability to infer. Since the AI Act does not clearly define what inference is, there is a gray area for certain data-driven systems. A specific example is credit scoring systems, which are listed by Annex III of the AI Act. At the same time, however, these are often implemented using statistical models for which it is unclear whether they have the capability to infer and thus fall under the AI definition of the AI Act at all. 

Motivated by statistical learning theory, this work develops a framework for grading different levels of the  capability to infer. Based on the AI Act and the  Commission Guidelines on the definition of an artificial intelligence system, we analyze which levels constitute sufficient capability to infer within the meaning of the AI Act and where further regulatory clarity is needed. We illustrate the framework by creating two realistic credit scoring workflows and show whether and where inference occurs in them. Our analysis illustrates that not only individual models but the entire data processing workflow must be considered. It also shows that the involvement of human experts during development can have significant influence on the capability to infer. Code can be found at \url{https://github.com/fraunhofer-iais/inference-framework-credit-scorecards}.  

\end{abstract}

\ccsdesc[500]{Social and professional topics~Governmental regulations}

\keywords{AI regulation, AI Act, AI definition, Inference, Credit Scoring}


\maketitle

\begin{table}[h]
\centering
\begin{tabular}{|r|c||c|c|}
\hline
\multicolumn{2}{|c||}{\textbf{Monthly Income}} & \multicolumn{2}{|c|}{\textbf{\makecell{Number of times \\ 30-59 days past due}}} \\
\hline
\textbf{Value} & \textbf{Points} & \textbf{Value} & \textbf{Points} \\
\hline
missing value   & 2 & 0 & 8 \\
$$[0, 3000]$$    & -4  & 1 & -13 \\
(3000, 5000]   & -3 & $\geq$ 2 & -29 \\
(5000, 7000]   & 0 &  &  \\
(7000, 10000]  & 3  & & \\
$\geq$ 10000  & 5 & & \\
\hline
\end{tabular}
\caption*{Scorecards like this can have a huge impact on customers. Are they a result of artificial intelligence as defined by the European AI Act?}
\label{tab:scorecard_debtratio}
\end{table}
\section{Introduction}

The European AI Act is the first comprehensive legal framework for artificial intelligence (AI) to come into force, subjecting high-risk systems in particular to strict regulatory requirements \cite{EU_AI_act}. The question of what constitutes an AI system is therefore central to the scope of this legal framework.

The AI Act opts for a technology-neutral definition that closely follows international preparatory work, particularly that of the OECD \cite{OECDRecommendations2024}, and formulates the capability to infer as the key distinguishing feature of AI systems. The inference of the AI system refers to the capability “to derive from received inputs for explicit or implicit goals how outputs such as predictions, recommendations, or decisions are generated.” However, despite its regulatory importance, the concept of inference remains under-specified, particularly with respect to systems that
rely on classical statistical models rather than contemporary machine learning architectures. Important examples for such statistical models are linear and logistic regression models. 

In practice, the discrepancy is particularly evident in credit scoring workflows: Annex III of the AI Act lists credit scoring. This implies that credit scoring systems are classified as high-risk systems by the AI Act, provided that the criteria outlined by Article 6 are met. However, their implementation is often based on (partially) automated binning procedures and logistic regression models. During the European Commission's consultation process on the definition of AI \cite{ConsultationAIdefinition}, industry representatives repeatedly expressed doubts as to whether logistic regression models should actually be considered AI systems \cite{AccisFeedback}. Although the European Commission published (legally non-binding) guidelines at the beginning of 2025 that further clarify the definition of AI \cite{AI_def_guidelines}, there is still uncertainty among industry practitioners and regulators as to whether logistic regression is considered AI within the meaning of the AI Act \cite{ECBNewsletter}. 

This ambiguity is also evident in the history of the AI Act: the first draft of the AI Act \cite{EU_AI_act} added an Annex to the AI definition, listing specific methods that were considered AI. In addition to machine learning, general statistical procedures and optimization methods were also mentioned. This approach was primarily criticized because it would have also covered many conventional software programs. In a compromise draft by the European Council \cite{EU_AI_Council}, this Annex was removed and replaced by two recitals referring to machine learning and logic- and knowledge-based approaches, listing logistic regression as a machine learning technique. 

The final version of the AI Act uses the technology-neutral definition as described above.

This work addresses this regulatory and conceptual ambiguity in two steps: First, we develop a framework for grading different levels of the  capability to infer. Based on the AI Act and the Commission Guidelines on the definition of an artificial intelligence system, we analyze which levels constitute sufficient capability to infer within the meaning of the AI Act and where further regulatory clarity is needed. We illustrate the framework by creating two realistic credit scoring workflows and analyze where and to what extent inference occurs within the workflows. Our analysis illustrates that not only individual models but the entire data processing workflow must be considered and shows that the involvement of human experts during development can have significant influence on the capability to infer.

\section{Related Work}

\subsubsection*{AI definition of the AI Act:} The discussion of the definition has primarily been viewed from the perspective of its genesis and the problem of formally defining the subject matter of regulation in legal terms.
\cite{Schuett02012023} argues that AI regulation in general should not rely on a definition of AI, as most existing AI definitions do not meet the requirements for legal definitions. 
\cite{Ebers2021} and  \cite{VealeZuiderveenBorgesius+2021+97+112} 
analyze the first proposal of the European Commission for the AI Act pointing out that the definition therein is too broad. \cite{Finocchiaro} considers that having a list of AI-techniques included in the AI Act may risk excluding future technological developments and \cite{Ellul} questions the feasibility of updating the list at a sufficient pace.
\cite{Castan2024}, \cite{Linera02092025} and \cite{Fernandez-LlorcaForthcoming-FERAIA-5} describe the process by which the definition was developed during the negotiations of the AI Act. The latter one also analyzes terms like AI system, generative AI etc. from an interdisciplinary perspective. \cite{Floridi2023} traces the development of the AI Act's definition of AI and examines the extent to which it is compatible with that of the American Executive Order. 
 \cite{Hacker2024} comments on the final trilogue version of the AI Act stating that inference is the only characteristic to distinguish AI systems from those built on classical software.

\subsubsection*{Credit scoring under the AI Act:} The regulation of credit scoring through the AI Act has been investigated through different lenses: \cite{Spindler02102021} and \cite{MONTAGNANI2024105984} analyze the impact of the (first draft version of the) AI Act regarding credit scoring, considering also existing banking regulation and its overlaps with the proposed AI Act. 
\cite{Hacker2025Future} examine the regulation of credit scoring in the context of underwriting and the regulatory landscape for insurance. 
Other work examines the legal requirements of individual trustworthiness dimensions for high-risk AI systems: \cite{Pavlidis02012024} investigates the AI Act's requirements for explainability. \cite{Floridi2026} propose a regulatory taxonomy that distinguishes transparency, traceability, interpretability, and explainability as layered and interdependent dimensions of AI opacity and exemplify these for credit scoring.

\subsubsection*{Different stages of inference:} 

In \cite{Breiman2001}, Breiman highlights a qualitative shift from parametric estimation to data-driven construction of decision logic. Decision tree learning explicitly constructs input-output mappings from data \cite{10.1023/A:1022643204877, books/wa/BreimanFOS84}, while instance-based and kernel methods realize inference through data-induced similarity geometries rather than explicit rules \cite{journals/tit/CoverH67, 10.1145/130385.130401}. Representation learning extends this dependence to the feature space itself, learning representations jointly with decision functions \cite{10.1109/TPAMI.2013.50, 10.1007/978-3-642-42051-1_16}.\\

While there is a lot of work on the AI Act, credit scoring, and different inference properties in the context of statistical learning, to the best of our knowledge, there is no work that analyzes the inference concept of the AI Act in greater technical detail.

\section{Analysis of the term inference}

In general, inference refers to "the drawing of a conclusion from known or assumed facts or statements" \cite{Inferencedefinition}. Statistical inference describes the process of drawing conclusions about a population based on data drawn from a sample of that population\footnote{The term population here refers to a set of similar items or events which is of interest for some question to be investigated.} \cite{casella2002statistical}. In machine learning, inference refers to the application of a trained model to a new data point. Since the AI Act is a legal text, the term inference must be interpreted strictly in its meaning. 
At the same time, the legal interpretation of this term needs to be operationalized by computer science.

Article 3 of the AI Act defines an AI system as “a machine-based system that is designed to operate with varying levels of autonomy and that may exhibit adaptiveness after deployment and that, for explicit or implicit objectives, infers, from the input it receives, how to generate outputs such as predictions, content, recommendations, or decisions that can influence physical or virtual environments.” As emphasized by Hacker \cite{Hacker2024}, the capability to infer is the decisive distinguishing feature, since the remaining elements of the definition may also be satisfied by conventional software.

Recital 12 clarifies that the "capability to infer refers to the process of obtaining the outputs, such as predictions, content, recommendations, or decisions, which can influence physical and virtual environments, and to a capability of AI systems to derive models or algorithms, or both, from inputs or data".  It emphasizes that the techniques  that enable inference include machine learning approaches that learn from data how to achieve certain objectives, while systems that rely exclusively on human-made rules or perform only simple data processing are excluded. The Commission’s legally non-binding Guidelines on the definition of an AI system further specify that this derivation primarily concerns the development phase of the system, without excluding the operational phase \cite{AI_def_guidelines}.

Taken together, Article 3 and Recital 12 establish inference as a structural criterion rather than a purely functional one. A system exhibits the capability to infer when the form of the input-output mapping that generates predictions, content, recommendations, or decisions is at least partially shaped by data, rather than being fully specified ex ante by human developers. However, the AI Act does not specify how strong this data-driven determination must be, leaving open where exactly the threshold for the capability to infer is to be drawn.

To systematize this issue and relate it to the threshold implied by the AI Act, we develop a framework\footnote{Our framework applies to learning systems, i.e. it does not address the "logic- and knowledge-based approaches" mentioned in Recital 12 of the AI Act.}  that distinguishes different levels of data involvement in shaping input-output mappings. As a conceptual starting point, we draw on the formal abstraction of learning problems in statistical learning theory \cite{mitchell1997machine}.

In supervised learning, a learning problem is characterized by an input space $X \subseteq \mathbb{R}^n$, an output space $Y$, and a generally not directly observable mechanism that assigns outputs to inputs. Inputs $x \in X$ are assumed to be drawn independently from an underlying probability distribution over $X$, and each observed output $y \in Y$ reflects the response of this unobservable mechanism to the corresponding input. The components of an input vector $x \in X$ are individual measurable input variables called features.
A learning system observes a finite data set $D = \{(x_i, y_i)\}_{i=0}^m$ and aims to construct a function $h: X \rightarrow Y$, i.e., a mapping from inputs to outputs, that generalizes beyond the observed sample $D$. The space of all such functions the learning system is allowed to choose from is called the hypothesis space\footnote{The so-called VC dimension \cite{Vapnik:71} provides a measure of how "large" this hypothesis space is. The "larger" the hypothesis space, the more learning options the associated system has; in other words, the greater its capability to infer. However, since calculating the VC dimension can be quite complicated, we have decided not to include it in the framework developed below for reasons of practical applicability.}. 
To assess the quality of such a function $h$, a loss function $l: Y \times Y \rightarrow \mathbb{R}_+$
is specified, which measures the discrepancy between predicted outputs and observed responses. Since the precise form of the data-generating process is not observable, learning algorithms rely on empirical criteria computed from the observed data to guide the construction of $h$. A specific and fully specified such function (including its structure and parameter values) is referred to as a model. The process of obtaining such a model from a given data set $D$ is called training. 

This abstraction generalizes beyond supervised learning. In unsupervised learning, only inputs $x_i \in X$ are observed, and learning aims at uncovering structural regularities in the distribution over $X$, such as clusters or latent representations, i.e., relations of similarity between inputs. 
In reinforcement learning, learning proceeds through sequential interaction with an environment, and decision functions are optimized with respect to cumulative reward signals rather than fixed target outputs. Despite these differences, all three paradigms share a common structural perspective: learning involves identifying admissible input-output mappings on the basis of data.

Crucially, data can affect admissible input-output mappings in qualitatively different ways. They may merely fix numerical parameters within a fully specified functional form, restrict the set of admissible structures through selection or regularization, or actively construct a novel logic according to which output is generated. 

In many cases, this influence is mediated by transformations of the input space. 
Feature selection \cite{10.5555/944919.944968} restricts admissible inputs, feature generation applies predefined transformations, and feature learning constructs representations, i.e., encodings of the original features in another space \cite{10.1109/TPAMI.2013.50, 10.1007/978-3-642-42051-1_16}, that are themselves data-dependent, thereby reshaping the space in which functions and output-generating structures are formed.

\subsection{A Staged Framework of Inference}

Building on this structural perspective, we propose a staged framework that distinguishes inference mechanisms according to how data influence the formation of input-output mappings. The framework is inspired by Breiman’s distinction \cite{Breiman2001} between the data modeling culture, in which the functional form of the input-output mapping is specified ex ante and data serve primarily to estimate parameters, and the algorithmic modeling culture, in which the logic of the input-output mapping itself is derived from data .

We refine this distinction by defining a sequence of inference mechanisms, each strictly extending the previous one: each level introduces an additional way in which data can shape input-output mappings. The five levels are summarized in Table \ref{tab:inference-framework}.

\subsubsection{Level 0: Fixed mapping}

At the lowest level, systems implement a fully specified mapping from inputs to outputs, defined entirely ex ante by human designers. There are no alternative input-output mappings available to the system. Data may be processed or filtered, but they do not influence parameters or structure. Such systems correspond to rule-based execution or basic data processing as described in Recital 12 of the AI Act.

\subsubsection{Level 1: Parametric adaptation within a fixed structure}

At the first non-trivial level, the functional form of the input-output mapping is fixed in advance, but contains free numerical parameters. Data are used to determine these parameter values, while the structure of the input-output mapping remains unchanged. 
Classical linear and logistic regression exemplify this setting. For instance, input-output mappings may be restricted to quadratic polynomials of the form $h(x) =a x^2+bx+c;\,\, a,b,c \in \mathbb{R}$, 
where learning consists solely in estimating the coefficients. Data thus determine a point within a predefined family of input-output mappings, but do not alter the functional form of the mapping itself. 

\subsubsection{Level 2: Structural selection}

At Level 2, data are used to select among a finite or discretely defined set of admissible input-output mappings. The available alternatives are specified ex ante, but data determine which one is employed.

A canonical example is stepwise feature selection. In its forward variant, the learning procedure starts from a minimal model containing no explanatory variables and successively adds features from a predefined pool. 

$L_1$-regularization also falls into this category. While implemented via continuous optimization, the constraint induces sharp exclusions of entire parameter combinations, which effectively removes structural alternatives from consideration.

\subsubsection{Level 3: Data-driven structural construction}

Beyond selection lies structural construction: input-output mappings are not chosen from a predefined catalog but are functionally constructed from data. The concrete form of the input-output mapping  is not specified ex ante. This can happen in an explicit (symbolic input-output mapping is learned) or an implicit (input-output mapping cannot be made explicit) manner. 

\subsubsection*{Level 3a: Explicit structural construction}

Decision trees and rule learning are canonical examples. While learning algorithms impose general constraints (e.g. admissible split operations or maximum depth), the concrete structure - splits, rules, and their arrangement - is built in response to the observed data. Data thus determine not only which structure is used, but how it is formed. 
This represents a stronger form of inference, as data induce novel structural configurations rather than merely activating predefined ones.

\subsubsection*{Level 3b: Implicit structural construction}

Structural construction need not yield an explicit symbolic model. Instance-based methods such as k-nearest neighbors or kernel methods rely on data-induced relational structures, such as neighborhood relations or similarity geometries. Here, the input-output mapping is realized through the geometry of the data rather than through explicitly represented rules.

\subsubsection{Level 4: Representational construction}
The strongest form of inference arises when data determine not only the input-output mapping but also the representational space in which the outputs are formed. In representation learning, features are no longer fixed inputs but learned objects. Deep neural networks and large language models exemplify this regime. Here, admissible input-output mappings are themselves reshaped by data through the learning of latent representations. \\

This framework does not by itself determine at which level the capability to infer is reached within the meaning of the AI Act. Rather, it provides a structured basis for interpreting this threshold and to make sure that similar borderline cases are treated in a similar manner. Indications for where the line may be drawn can be derived from the (legally non-binding) Commission Guidelines on the definition of an artificial intelligence system \cite{AI_def_guidelines}. However, as their explanations are a bit confusing, their interpretation is not straightforward. Article 42 of these guidelines states that "systems used to improve mathematical optimization or to accelerate and approximate traditional, well established optimization methods, such as linear or logistic regression methods, fall outside the scope of the AI system definition. This is because, while those models have the capacity to infer, they do not transcend ‘basic data processing’." 
The confusion arises for two reasons: First, Recital 12 of the AI Act states that "the capacity of an AI system to infer \textit{transcends} basic data processing", i.e., the capacity to infer \textit{is more} than basic data processing. Second, one can view the construction of regression models - just like any other machine learning model as well - as the application of a mathematical optimization procedure, namely the minimization of the loss function. However, there is generally no connection where the improvement, acceleration or approximation of optimization methods plays a role in regression models (nor in machine learning models), which causes additional confusion.   

If Article 42 is interpreted to mean that systems based on regression models do not constitute AI in the sense of AI Act, the threshold for the capability to infer must be set as follows: systems at Level 1 may fall outside the scope of the AI system definition, as their structure is fully specified ex ante and data merely determine parameters. By contrast, systems at Level 3 and above closely align with the notion of “deriving models or algorithms from data” in Recital 12, as the input-output mapping is functionally constructed from data. Systems at Level 2 occupy an intermediate position, since they involve data-driven selection among predefined structures without fully constructing them.

At the same time, the ambiguity of the Guidelines leaves room for an alternative interpretation: one could argue that even parameter estimation within a fixed functional form already constitutes a form of inference, as the model is fitted to data and thereby “derived” from it. On this reading, the capability to infer could already be present at Level 1, shifting the threshold of the AI system definition accordingly.

As the Commission Guidelines \cite{AI_def_guidelines} emphasize that “the notions of autonomy and inference go hand in hand”, we finally analyze the relationship between autonomy and the capability to infer. Our discussion shows that the key distinguishing feature of AI systems is the capability to infer and that autonomy plays only a secondary role: In classical AI literature, autonomy is often understood in epistemic terms. For instance, \cite{russell2021} define autonomy as the extent to which an agent\footnote{Note that this term refers to the notion of "classical" AI agents in the computer science literature, however it also holds true for "modern", LLM-based AI-agents.} relies on its own percepts and learning processes rather than on prior knowledge provided by its designer. On this understanding, autonomy is closely linked to the system’s ability to learn from and adapt to data, and thus conceptually closely aligned with what the AI Act describes as the capability to infer.
However, the concept of autonomy introduced by the AI Act is more reminiscent of automation: Article 3 of the AI Act requires that a system is “designed to operate with varying levels of autonomy”, which Recital 12 defines as some degree of independence from human involvement and the capabilities to operate without human intervention. This particular requirement would also be fulfilled by systems that are based on "classical" software (i.e. systems that are not considered to be AI according to Recital 12). The Guidelines  refine the concept somewhat: They further state that a "system that requires manually provided inputs to generate an output by itself is a system with ‘some degree of independence of action’, because the system is designed with the capability to generate an output without this output being manually controlled, or explicitly and exactly specified by a human." If one interprets the last half-sentence to mean that systems based on the rules defined solely by natural persons do not have autonomy, one is again close to the computer science concept of autonomy, which in turn is closely linked to the concept of the capability to infer in the AI Act. Otherwise, if the concept of autonomy is  interpreted in the meaning of automation, it cannot serve as a distinguishing feature of AI systems.

\begin{table*}[ht]
\centering
\small
\begin{tabular}{p{0.6cm} p{3.8cm} p{3.8cm} p{6.5cm}}
\hline
\textbf{Level} &
\textbf{Inference mechanism} &
\textbf{Role of data in shaping the input-output mapping} &
\textbf{Examples} \\ 
\hline 
0 & Fixed mapping &
No restriction or adaptation of structure by data &
 Hard-coded rules, deterministic decision logic, fixed descriptive statistics (means, counts) \\ 

1 & Parametric adaptation &
Data determine parameter values within a fixed structure &
 Linear regression, logistic regression, $k$-means with fixed $k$, principal component analysis with fixed dimensionality \\ 

2 &
Structural selection &
Data select among predefined alternatives with fixed structure &
 Stepwise feature selection, L1-regularized regression, data-driven selection of number of clusters $k$ in $k$-means \\ 

3a &
Structural construction (explicit) &
Data construct symbolic or combinatorial structure &
 Decision trees, rule learning, 
hierarchical clustering, tree-based discretization \\ 

3b &
Structural construction (implicit) &
Data induce relational or geometric structure &
 $k$-nearest neighbors, kernel support vector machines,  kernel principal component analysis \\ 

4 &
Representational construction &
Data induce the representational space itself & Deep neural networks, transformers, autoencoders, self-supervised representation learning \\ 

\end{tabular}
\caption{Staged framework of inference mechanisms that derive input-output mappings from data}
\label{tab:inference-framework}
\end{table*}

\subsection{Categorization of typical data driven models used for credit scoring}

In order to prepare our analysis of the credit scoring use case, we investigate binning and logistic regression with regard to our framework. Binning is commonly used as a data preprocessing technique rather than considered as a machine learning model on its own. Within the meaning of the AI Act, binning is not considered AI as it does not produce predictions, decisions, recommendations, or outputs. However, it can contribute to the overall capability to infer of an AI system.  

\subsubsection{Binning}
\label{sec:binning}

Binning transforms a numerical input variable into a categorical representation by partitioning its domain into intervals. In manual binning, interval boundaries are defined ex ante based on expert judgment or regulatory conventions. Similarly, quantile-based binning fixes the number of bins $K$ and defines intervals via empirical quantiles: $B_k = [Q_{(k-1)}, Q_{k}),$ where $Q_k$ denotes splitting points chosen so that each bin $B_k$ comprises the same number of observations. Although quantile binning uses data to determine numerical cut points, the structural form of the partition—equal-frequency bins with fixed cardinality—is fully specified in advance and independent of the target variable. Thus it remains below structural construction in the framework. 

By contrast, data-driven binning methods commonly used in credit scoring rely on recursive binary splits that optimize a target-dependent impurity criterion, most prominently the Shannon entropy. Let $\mathcal{X}$ be a set of instances with target variable $y$ taking values in ${Y}$. Then the Shannon entropy of the target variable in $\mathcal{X}$ is defined as
\begin{equation}
   H(\mathcal{X}) = - \sum_{y \in {Y}} p(y) \log p(y),
\end{equation}
where $p(y)$ is the proportion of class $y\in {Y}$ in $\mathcal{X}$ \cite{StatisticalLearning2009}. For a candidate split $s$ that partitions the set $\mathcal{X}$ into regions $R_1$ and $R_2$, the information gain $IG(s)$ is defined as 
\begin{equation}
    IG(s) = H(\mathcal{X}) - \sum_{k=1}^2 \frac{|R_k|}{|\mathcal{X}|} H(R_k).
\end{equation}
The algorithm recursively selects the split that maximizes $IG(s)$ subject to stopping criteria such as a maximum number of bins or a minimum relative reduction in entropy.

Although practical implementations impose explicit constraints, such as these stopping criteria, the resulting partition is not selected from a predefined catalog of admissible structures.\footnote{If one can enumerate the possible splitting points in advance and the number of splitting points is fixed then capability to infer might be reduced to Level 2. Consider e.g., the interval $[0,100]$. In case splits are only allowed at natural numbers and the number of total splits is $k$, then there are $\bigg(\begin{array}{c} 99 \\ k \end{array}\bigg)$ potential splits. Due to the fast growth of this number for $k \to 50$, the bins will effectively be determined by the data as, for example, the CART algorithm \cite{books/wa/BreimanFOS84} does not consider all splits but is searching for a local minimum. However, for small $k \sim 1,2$, the system does not fall into Level 3.} Instead, the input-output mapping—i.e., the exact number, location, and hierarchy of splits—is functionally constructed from the data via an inductive procedure. It is thus considered Level 3a.

\subsubsection{Logistic regression} \label{logistic regression}
Logistic regression assumes a fixed parametric form for the conditional probability of the target variable given an input vector $x \in \mathbb{R}^d$:
\begin{equation}
\mathbb{P}(y=1 \mid x) = \sigma(w^\top x + b),
\qquad \sigma(z) = \frac{1}{1 + e^{-z}}.
\end{equation}
Learning consists of estimating the parameter vector $w \in \mathbb{R}^d$ and bias term $b \in \mathbb{R}$, typically by minimizing a convex loss function such as the negative log-likelihood. The hypothesis space is fixed and finite-dimensional, and data influence only the numerical values of the parameters. In this configuration, logistic regression exemplifies Level 1 (parametric adaptation), as no structural aspects of the decision logic are derived from data. 

In practical credit-scoring applications, logistic regression models are often trained using stepwise feature selection \cite{Bursac, hosmer.lemeshow:applied}. Here, in the forward variant, one starts with a model with the "best" variable where "best" is determined using pre-determined criteria, such as the Wald test for statistical significance of that variable. Successively, further variables are added until a stopping condition is reached. 

While the functional form of the model remains unchanged, data determine which structural alternative—namely, which feature subset—is selected. Such configurations therefore correspond to Level~2 (structural selection), as inference operates through data-driven choice among predefined structural options rather than through structural construction.

Taken together, these examples illustrate that credit-scoring systems frequently combine components operating at different levels of inferential capability. Analyzing such systems through the proposed framework allows their constituent techniques to be disentangled and supports a more granular assessment of which elements plausibly instantiate the capability to infer as described in Article~3 and Recital~12 of the AI Act.

\section{Capability to infer of creditworthiness models}

In order to analyze whether statistical models used for credit scoring have the capability to infer and thus fall under the AI definition of the AI Act, we apply the framework developed in the last section to two examples of such credit scoring models. We start by recalling how such credit scoring models are developed for typical industrial applications.

\subsection{Classical scorecard development workflow}
\label{sec:scorecard_pipeline}

Classical credit scorecards remain a widely deployed and regulatorily accepted approach for assessing creditworthiness, particularly in retail lending and consumer finance \cite{Siddiqi2006, Buecker2022}. Despite their apparent simplicity, scorecards are typically developed through a structured, multi-stage process that combines statistical modeling with domain expertise and governance constraints \cite{Siddiqi2006, Buecker2022}.

The process begins with the definition of the target variable and associated observation and performance windows. In credit risk applications, this usually entails a binary default indicator, such as a 90-days-past-due event within a fixed horizon \cite{HandHenley1997, Siddiqi2006, Anderson2007}. Careful specification of these elements is critical, as it determines both the semantic meaning of the model output and its regulatory validity.

Following target definition, the raw dataset, often comprising hundreds of candidate variables, is subjected to extensive data cleaning and preprocessing. This includes consistency checks, treatment of missing values, and outlier handling, with the objective of ensuring robustness and stability rather than maximizing predictive performance \cite{Finlay2012}. At this stage, variables may be removed for purely technical or regulatory reasons, without reference to predictive power.

A subsequent variable shortlisting phase reduces the dimensionality of the dataset. This step typically combines domain knowledge, regulatory constraints, and simple univariate statistics such as missing value rates or stability measures. While quantitative indicators may be used, the shortlisting process in practice often remains partially judgment-driven \cite{Siddiqi2006}.

Continuous and ordinal variables are then discretized through binning. Binning serves multiple purposes: it stabilizes variable behavior over time, facilitates interpretability, and prepares the data for transformation into Weight-of-Evidence (WoE) values (as defined in the next paragraph). Common approaches include manual binning as well as tree-based binning methods, with missing values typically treated as separate bins \cite{Siddiqi2006}. Typical loss functions for tree-based methods are the Gini Impurity or the Shannon entropy and Information Gain, as defined in Section \ref{sec:binning}.

A common constraint considered in the binning process is that every feature should have a clear direction of effect, i.e., the WoE values for non-missing values should monotonically increase or monotonically decrease across the bins, without any reversals \cite{Siddiqi2006, Finlay2012}. This ensures that only features with comprehensible effects on the default predictions are used in scoring, in the interest of transparency. Further common constraints are that each bin shall contain a minimum of 5\% of instances, and that there shall not be any bins with 0 counts of 'bad' or 'good' cases \cite{Siddiqi2006}. Note that while these constraints are not regulatory requirements, they are well-established best practices.
After binning, further variables may be excluded due to lack of adherence to these constraints or any others previously set for the process.

The binned variables are transformed using the Weight of Evidence (WoE) framework \cite{HandHenley1997}, which encodes each bin in terms of its ability to distinguish between non-defaults and defaults. This transformation linearizes the relationship between predictors and the log-odds of default, enabling the effective use of logistic regression while preserving interpretability.

Univariate measures such as the Information Value (IV) are then employed to assess the predictive contribution of individual variables \cite{Siddiqi2006}. Variables with low discriminatory power or poor stability are removed prior to multivariate modeling.

A typical approach to estimating the core model is using logistic regression on the WoE-transformed variables \cite{Siddiqi2006, Anderson2007, Finlay2012, Lessmann2015, Buecker2022}. Variable selection may be performed using stepwise procedures, subject to constraints on multicollinearity and model complexity. Common metrics for model performance used in variable selection include, among others, the Area under the Receiver Operating Curve (AUROC), the Gini coefficient, and the Average Precision score (AP) \cite{Siddiqi2006, DavisGoadrich2006}. Formal definitions of WoE, IV, AUROC, Gini, and AP are provided in Appendix \ref{app:definitions}.

While AUROC and Gini are sensitive to skewed distributions of the target variable, and can produce high scores in spite of poor performance on the minority class, AP is more robust to such imbalances \cite{DavisGoadrich2006}. In credit scoring, distributions of non-default and default tend to be highly skewed towards non-default \cite{Siddiqi2006, Finlay2012} and thus, indicators that are robust to class imbalance should be considered in both the construction and evaluation of scoring models.
Multicollinearity can be controlled using, for example, the Variance Inflation Factor (VIF) to identify highly correlated variables \cite{Finlay2012}.

The resulting final model trained on the selected variables is then validated using out-of-sample testing and stability metrics to ensure robustness over time. 

Finally, the model is converted into an operational scorecard by defining a score offset and performing points-to-double-the-odds (PDO) score scaling, in which the points associated with the feature bins are scaled such that an increase of a certain number of points in the score corresponds to a doubling of the odds of non-default \cite{Siddiqi2006}. The completed scorecard is embedded within a governance framework that includes documentation, approval processes, monitoring, and periodic review, in line with regulatory expectations.

\subsection{Illustration using two credit scoring examples}
\label{sec:cs_examples}

We now illustrate our framework using two example credit scoring approaches that exhibit different degrees of inference capability. Both examples are built using the Give Me Some Credit dataset from Kaggle \cite{GiveMeSomeCredit}, which contains 150.000 instances with 10 features and the binary target variable of a default to be predicted. A default here means a delinquency of over 90 days past due. Definitions of all features as well as the full resulting scorecards are presented in Appendix \ref{app:cs_inference}. 

Before constructing the two models, we perform preprocessing of the data as follows. Missing values for \textit{MoInc} are marked for treatment as a separate bin later on. Neither the target variable nor any of the other features have missing values for any of the instances in the dataset. Furthermore, we want to exclude the features \textit{Age} and \textit{NumberOfDependents} from use for scoring for ethical reasons and thus discard them from the data. Lastly, the data is split into 70\% training and 30\% test data.

The binning procedures differ between the models, but both consider the best practice constraints as stated in Section \ref{sec:scorecard_pipeline}: Each bin must contain a minimum of 5\% of instances, there must not be any bins with 0 counts of default or non-default cases respectively, and we enforce monotonicity of the WoE values across bins for every feature, with exception of the missing value bin of \textit{MoInc} as well as for the 0 values of selected features, based on domain-knowledge. In preparation for binning of the features, we thus consider the direction of that monotonicity for every feature. For example, a higher \textit{MoInc} points to a higher probability of on-time payments and thus a lower risk of default, whereas for the feature \textit{N30-59Late}, which counts the number of times that a person has already had payments be 30 to 59 days past due previously in their credit history, the higher that number, the higher the risk of future default may be.\footnote{Definitions of all features and their respective expected directions of monotonicity, including the rationale behind them, are presented in Appendix \ref{app:cs_inference}}.

Feature selection processes differ strongly between approaches and are discussed individually in the respective following sections.

The models themselves are both constructed with logistic regression, using the scikit-learn implementation \texttt{LogisticRegression} (with the 'liblinear' solver). Lastly, scorecards are constructed using scaling with PDO, setting $20$ as the number of points to double the odds of a non-default outcome and a score-offset of $600$ points. The final scorecards for both models are given in Appendix \ref{app:cs_inference}.

\subsubsection{Example I: Semi-automated credit scorecard construction}

\begin{figure}[htb]
    \centering
    \includegraphics[width=\columnwidth]{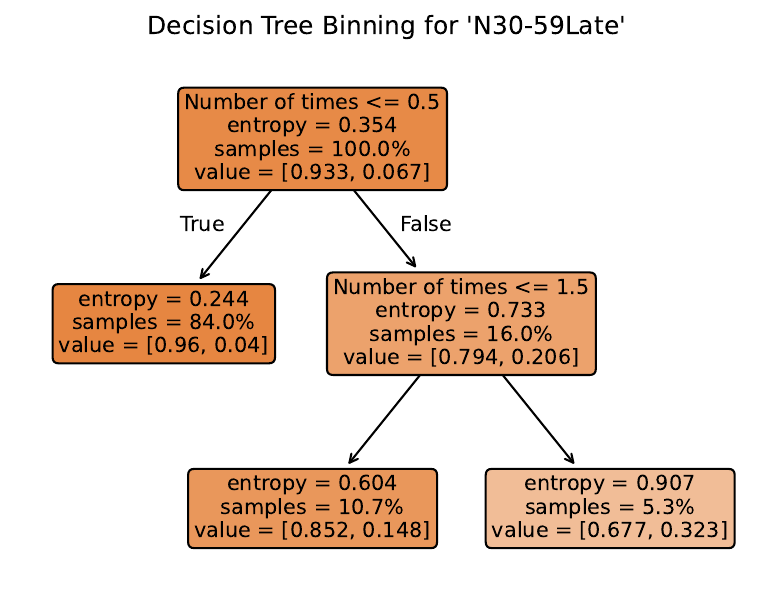}
    \caption{Decision tree binning for the feature \textit{N30-59Late}. Node labels show split criteria, and (rounded) sample proportions, class distributions in the format \texttt{value = [good, bad]}, and entropy values for each bin. Note that this is a simple tree with only three bins for illustration, and more complex trees are constructed in Example I.} 
    \label{fig:tree_30-59}
\end{figure}

After data preprocessing as outlined above, binning of the feature variables is performed using the \texttt{DecisionTreeClassifier} from scikit-learn, which, node by node, determines the split point that best separates default and non-default cases, optimizing for minimal entropy in each bin. The monotonic constraints are passed to the \texttt{DecisionTreeClassifier} encoded as values of 1 and -1 for monotonic increase or decrease, respectively, for each feature. Based on these constraints, the algorithm only allows splits where the average probability of default either monotonically increases or monotonically decreases across bins, automatically rejecting splits that would violate monotonicity. Similarly, the 5\% rule is passed to the function by setting the \texttt{min\_samples\_leaf} parameter to $0.05$, and the condition of no 0 counts of either default or non-default instances per bin is enforced via an if-condition in the surrounding algorithm. An example of a resulting decision tree is given in Figure \ref{fig:tree_30-59}. This method of binning is data-driven while simultaneously being consistent with domain expectations and the given constraints.

Next, we train a logistic regression model to statistically infer the relationship between the WoE-transformed features and the target variable. The selection of features for use in the final scorecard is performed stepwise, iteratively choosing the most predictive features in an automated process. At each step, the algorithm evaluates all remaining features and selects the one that warrants the largest improvement of model performance, as measured by the average precision score. If the selected feature improves the score by at least $0.002$, it is added to the model, otherwise, it is discarded and the process stops. At the end of every forward step, a backward check of multicollinearity is performed and any feature with a VIF greater than $10$ is removed from the model\footnote{Following a rule of thumb as stated in \cite{Finlay2012}.}. 

The final logistic regression model, trained on the WoE values and ground truths of the selected features, achieves an AP of 0.3737 and an AUROC of 0.8517.

\subsubsection{Example II: Manual credit scorecard construction}

After data preprocessing, we perform manual binning and feature selection largely following the process described in \cite{Siddiqi2006}. 
For each feature in the dataset, we first consider equal-frequency binnings and calculate Bad Rates (i.e., percentage of default outcomes) and WoE values per bin, as well as IV per binning. For further manual design of bin split points, these statistics are sorted into easy-to-comprehend tables, an example of one is given in Table \ref{tab:income}. Based on this information, we then manually and iteratively define split points to create improved binnings, with the aim of maximizing IV, under consideration of the aforestated best practice conditions. Features that cannot be split into bins with WoE values that adhere to all constraints while simultaneously being intuitively explainable are excluded from further consideration for the score model.

\begin{table}[htbp]
\centering
\begin{tabular}{|r|c|c|c|c|}
\hline
\multicolumn{5}{|c|}{\textbf{\textit{MoInc} (Monthly Income)} \hspace{1cm} IV = 0.0721} \\
\hline
\textbf{Bin} & \textbf{Count} & \textbf{Bads} & \textbf{Bad Rate} & \textbf{WoE} \\
\hline
missing value            & 20723 & 1136 & 5.48\% & 0.21 \\
$[0.0,\, 2692.0]$     & 14047 & 1236 & 8.80\% & -0.30 \\
$(2692.0,\, 4000.0]$     & 14316 & 1306 & 9.12\% & -0.34 \\
$(4000.0,\, 5382.0]$     & 13778 & 1071 & 7.77\% & -0.16 \\
$(5382.0,\, 7059.0]$     & 14045 & 898  & 6.39\% & 0.05 \\
$(7059.0,\, 9900.0]$     & 14092 & 758  & 5.38\% & 0.23 \\
$\geq$ 9900.0  & 13999 & 613  & 4.38\% & 0.45 \\
\hline
\end{tabular}
\caption{Equal-frequency binning overview for feature \textit{MoInc}.}
\label{tab:income}
\end{table}

Of the remaining features, all features with an IV of $0.02$ or greater are selected for use in the model. 
For the final set of selected features, WoE tables are calculated, and the final regression model is trained on the WoE values and the target variable ground truths. This process results in a model with an AP of 0.3602 and AUROC of 0.8502 on the test data.

\section{Inference levels of the semi-automated and manual credit scoring approaches}
Both scorecard construction approaches share the constraints of the binning, the use of logistic regression to train the final model, and the employed indicators of IV, WoE values, and average precision score. Yet, they differ fundamentally in the degree to which the construction of the model's input-output mapping is driven by the data.

Data preprocessing is performed identically for both approaches and without any data-driven construction of model structure, as the treatment of missing values and the exclusion of features for ethical reasons occur manually. 

The first significant conceptual difference lies in the binning process. In Example I, split points are determined by a decision tree optimizing entropy, subject to the human-defined constraints of monotonicity and minimum numbers of instances per bin. Here, the data actively construct the input-output mapping, corresponding to Level 3a in our framework. In contrast, Example II employs an iterative manual binning based on domain knowledge and quantitative indicators. The structure is determined entirely by human reasoning, corresponding to Level 0.

For Example II, the feature selection is performed manually based on the Information Value of the binned features. A logistic regression model is trained on these preselected variables, which represents a Level 1 case.
Example I employs an automated stepwise data-driven process to train a logistic model which, in every step, performs model training and selects the AP maximizing feature without active human intervention. This constitutes a Level 2 case. 

In summary, the first approach, due to its data-driven tree-based binning and the resulting data-driven construction of the model's input-output mapping, is overall classified as Level 3a (explicit structural construction) as defined in Table \ref{tab:inference-framework}. The second approach is based on manual design decisions throughout, but employs logistic regression for parameter fitting. Thus, it is classified as Level 1 (parametric adaptation). Here, we have tacitly introduced the rule that the data processing component with the highest inference level determines the inference level of the overall system.

\subsection{Resulting scorecards and the impact of human intervention on inference levels}
The complete scorecards are found in Appendix \ref{app:cs_inference}, an excerpt for purposes of comparison and discussion is presented in Table \ref{tab:scorecard_example_comparison}. 
For some features, the two approaches produce extremely similar results. For example, for \textit{N30-59Late} (number of minor prior delinquencies between 30 and 59 days), both approaches produce the same effective binnings and corresponding WoEs, with only minor deviations in points as a result of differences in the scorings of other attributes (see Table \ref{tab:scorecard_example_comparison}). Results are especially similar for features that only take integer values and thus offer significantly fewer candidate split points that still fulfill all constraints than real-valued features.

\begin{table}[htbp]
\centering
\small
\begin{tabular}{|rrr|rrr|}
\toprule
\multicolumn{3}{|c|}{\textbf{Example I (semi-automated)}} & \multicolumn{3}{c|}{\textbf{Example II (manual)}} \\
\midrule
\textbf{Bin} & \textbf{WoE} & \textbf{Pts} & \textbf{Bin} & \textbf{WoE} & \textbf{Pts} \\
\midrule
\multicolumn{6}{|c|}{\textit{\textbf{N30-59Late}}} \\
\midrule
 0 & 0.532 & 8 & 0 & 0.532 & 8  \\
 1 & -0.882 & -13 & 1 & -0.882 & -14 \\
$\geq$ 2 & -1.895 & -28 & $\geq$ 2 & -1.895 & -29 \\
\midrule\multicolumn{6}{|c|}{\textit{\textbf{MoInc}}} \\
\midrule
missing value & 0.211 & 3 & missing value & 0.211 & 2 \\
$$[0.0, 3332.5]$$ & -0.355 & -5 & [0.0, 3000.0] & -0.334 & -3 \\
(3332.5, 3920.5] & -0.246 & -3 & (3000.0, 5000.0] & -0.227 & -2  \\
(3920.5, 4621.5] & -0.220 & -3 &  (5000.0, 7000.0] & 0.009 & 0 \\
(4621.5, 5563.5] & -0.106 & -1 & (7000.0, 10000.0] & 0.231 & 2 \\
(5563.5, 6550.5] & 0.002 & 0 & $\geq$ 10000.0 & 0.464 & 5  \\
(6550.5, 7656.5] & 0.190 & 2 & & & \\
(7656.5, 10284.0] & 0.286 & 4 & & & \\
$\geq$ 10284.0 & 0.474 & 6 & & & \\
\bottomrule
\end{tabular}
\caption{Comparison of scorecards for attributes \textit{N30-59Late} and \textit{MoInc}, with bin split points, WoE values and points rounded.}
\label{tab:scorecard_example_comparison}
\end{table}

Comparing the scorecards for the monthly income feature \textit{MoInc} (see Table \ref{tab:scorecard_example_comparison}), the \texttt{DecisionTreeClassifier} binning created more splits. The second and third bin map to very similar WoE values and the exact same number of points (due to rounding effects). In practice, a human expert would likely decide to combine these two bins from the Example I scorecard into one. While this constitutes a human intervention, the basic structure of the binning as performed by the decision tree would still persist, and the binning process overall would remain classified as Level 3a in our framework. 

However, if a human expert completely overrules the data-derived split points and continues the scorecard design workflow with entirely human-defined binnings, such as adopting the \textit{MoInc} structure from Example II instead of Example I, the significance of the tree-based binning for the overall workflow is lost.

In summary, the degree and type of human intervention determine whether the binning retains its Level 3a classification. Minor adjustments, such as combining similar bins, merely constitute further processing of the data-derived structure and preserve the data-driven nature of the process. In contrast, complete human overruling of the data-derived binning, such as setting new split points, destroys the capability to infer.

\subsection{The need to consider complete workflows}
The assessment of the level of inferential capability that a system possesses requires examination of its entire workflow rather than isolated components. Consider the binning process from Example I in isolation: while it employs automated decision tree binning, this does not constitute an AI system on its own, as it does not produce predictions, decisions, recommendations, or outputs. Instead, this process is of structure-defining significance only when integrated into the credit scorecard modeling workflow, where it contributes to determining the ultimate overall inferential capability of the system in the sense of the AI Act. Vice versa, if one were to assess the capability to infer based solely on the final model — logistic regression — one would arrive at an incorrect result.

\section{Conclusion}
This work addresses the classification of the inferential capability of data-driven models within the meaning of the European AI Act. The AI Act formulates the capability to infer as a key distinguishing feature for the definition of artificial intelligence. At the same time, it provides only vague indications of the conditions under which inferential capability exists, making it unclear for certain borderline cases of data-driven models whether they have sufficient inferential capability to be considered AI and thus regulated by the AI Act. The framework we have developed builds on the fundamentals of statistical learning theory and defines five levels of inference: fixed mappings without any inference capability (Level 0), parametric models (Level 1), selection from predefined structural alternatives (Level 2), data-driven structural construction (Level 3), and models that learn the relevant features themselves (Level 4). Our framework cannot determine exactly where the threshold for the capability to infer lies, but it does help to systematize this discussion. In addition, it ensures comparability when assessing similar borderline cases.  While Level 0 is certainly not AI within the meaning of the AI Act and Levels 3 and 4 have the capability to infer, the classification of Levels 1 and 2 is unclear. The legally non-binding Commission Guidelines on the definition of an AI system might suggest that Level 1 should be classified as insufficient capability to infer. 

By applying the framework to a specific use case—credit scoring—we illustrate that the entire development data processing workflow must be considered when determining an AI system's capability to infer. Credit scoring represents an interesting use case because, on the one hand, it is listed by Annex III of the AI Act and is thus a high-risk system, provided the criteria as outlined by Article 6 are met. On the other hand, it typically uses techniques whose inference capabilities are unclear a priori. For our analysis, we constructed two realistic credit score cards based on the Give Me Some Data dataset, both of which are based on logistic regression. The first uses data-driven methods for binning and feature selection for logistic regression and is classified as Level 3 in our framework. The second is based on manual binning and feature selection and is classified as a parametric model in Level 1. Furthermore, we show that the capability to infer can be lost when human interventions destroy the data-inferred input-output mapping.



\section{Limitations}
The dataset we use has fewer instances and features than a typical one used in the industry. However, this does not limit our analysis, as it aims to systematically evaluate the inference capability of an end-to-end credit scoring workflow in \ref{sec:cs_examples}.

The framework primarily addresses AI systems that are developed in a data-driven manner. An extension to symbolic AI systems would be interesting.

\section{Statement of generative AI use}
The authors used generative AI tools (GPT 5, GPT 4, DeepL, Qwen3 Coder) to set LaTeX tables and as feedback tools for code and text improvement. Most of the text and code are human-written, but individual sentences and code cells were generated using these tools. The authors reviewed and are responsible for all content.

\section{Acknowledgments}
This work was supported by the Ministry of Economic Affairs, Industry, Climate Action and Energy of the State of North Rhine-Westphalia as part of the flagship project ZERTIFIZIERTE KI. 


\bibliographystyle{ACM-Reference-Format}
\bibliography{cleaned}

\appendix

\section{Definitions for scorecard development}
\label{app:definitions}
The credit scorecard development workflow as laid out in Section \ref{sec:scorecard_pipeline} utilizes several values encoding the relationship between default and non-default data points, in an individual bin (Weight of Evidence) or a set of bins (Information Value), as well as metrics for the quality of a model (AUROC, Gini coefficient, Average Precision).
They are defined as follows.

\textbf{}
\begin{definition}[Weight of Evidence (WoE)]
The WoE framework expresses each bin as the logarithm of the ratio between non-default and default frequencies as follows: For a bin $B$, the WoE is \cite{HandHenley1997}
\begin{equation}
    WoE(B) = \ln p_0^B - \ln p_1^B .
\end{equation}
\end{definition} 

\begin{definition}[Information Value (IV)]
The IV of a feature $F$ is calculated as 
\begin{equation*}
    IV(F) = \sum_{B \in \text{Bins}_F} (p_0^B - p_1^B) \left(\ln\left(p_0^B\right) -\ln\left(p_1^B\right)\right),
\end{equation*}
where $p_0^B$ and $p_1^B$ are the proportions of 'good' (non-default) and 'bad' (default) outcomes in bin $B$, respectively \cite{Siddiqi2006}.
\end{definition}

\begin{definition}[Area under the Receiver Operating Characteristic Curve (AUROC)]
For a model $m$, $AUROC(m)$ is the area under the curve that plots the False Positive Rate of $m$ against its True Positive Rate, and takes values in $[0, 1]$. A higher AUROC signifies better performance \cite{Siddiqi2006, DavisGoadrich2006}. 
\end{definition}

\begin{definition}[Gini coefficient]
The Gini coefficient of a model $m$ is equivalent to the AUROC via $Gini(m) = 2\cdot AUROC(m) - 1$, with values in $[-1, 1]$ and a higher value again signifying better performance \cite{Siddiqi2006, DavisGoadrich2006}. 
\end{definition}

\begin{definition}[Average Precision (AP)]
The AP of model $m$ corresponds to the area under the curve plotting its Precision against its Recall and takes values in $[0,1]$, the higher the better the performance \cite{Siddiqi2006, DavisGoadrich2006}. 
\end{definition}

Note that the definitions of Shannon entropy and Information Gain, which are commonly used as loss functions for decision tree-based binning methods, are already included in Section \ref{sec:binning} and thus are not repeated here.

\section{Credit scorecard examples}
\label{app:cs_inference}
The Give Me Some Credit Kaggle competition \cite{GiveMeSomeCredit} provides separate files for training and test data. Since the latter does not include the ground truth, we use and split the training data file exclusively. It contains 150,000 instances with 11 features. 

For each of the features, a respective monotonicity constraint was applied in our examples in Section \ref{sec:cs_examples}. The definitions of the features and the rationales behind their expected directions of monotonicity are as follows:

\begin{itemize}
    \item \textbf{SeriousDlqin2yrs (binary):} Person experienced 90 days past due delinquency or worse. (Target variable)
\\
    \item \textbf{MonthlyIncome (short name \textit{MoInc}, real):} Monthly income. The higher the income, the more money is available, the lower the risk of default. Missing values treated as separate bin.
\\
    \item \textbf{RevolvingUtilizationOfUnsecuredLines (\textit{RevUtil}, real):} The borrower's total balance on credit cards and personal lines of credit, except real estate and installment debt like car loans, divided by the sum of credit limits. The higher the value, the higher the risk of default.
\\
    \item \textbf{DebtRatio (real):} The borrower's monthly debt payments, alimony, and living costs, divided by monthly gross income. The higher the value, the less income available, thus, the higher the risk of default.
\\
    \item \textbf{NumberOfOpenCreditLinesAndLoans \\ (\textit{NumLinesLoans}, integer):} Number of open loans and lines of credit (e.g.\ credit cards). The more open loans, the higher the risk of default, with value 0 excepted from monotonicity.
\\    
    \item \textbf{NumberRealEstateLoansOrLines (\textit{NumRealEstate}, integer):} Number of mortgage and real estate loans including home equity lines of credit. The more open loans, the higher the risk of default, with value 0 excepted from monotonicity.
\\
    \item \textbf{NumberOfTime30-59DaysPastDueNotWorse \\ (\textit{N30-59Late}, integer):} Number of times borrower has been 30--59 days past due in the last 2 years. The higher the number of previous delays, the higher the risk of default.
\\
    \item \textbf{NumberOfTime60-89DaysPastDueNotWorse \\ (\textit{N60-89Late}, integer):} Number of times borrower has been 60--89 days past due in the last 2 years. The higher the number of previous delays, the higher the risk of default.
\\
    \item \textbf{NumberOfTimes90DaysLate (\textit{N90Late},integer):} Number of times borrower has been 90 days or more past due before. The higher the number of previous defaults, the higher the risk of default.
\\
    \item \textbf{Age (integer):} Age of the borrower in years. Excluded for fairness.
\\
    \item \textbf{NumberOfDependents (integer):} Number of dependents in family excluding themselves (spouse, children etc.). Excluded for fairness.
\end{itemize}
Excepting the value 0 from the monotonicity requirement for certain features, as is the case here for \textit{NumLinesLoans} and \textit{NumRealEstate}, is common practice in credit scoring. This is due to the fact that less information and a shorter (if any) repayment history may be available for applicants with no loans or lines, and very low utilization accounts can demonstrate higher risk \cite{Siddiqi2006}.


The complete scorecards resulting from Examples I and II are presented in the following tables. They offer a direct comparison of the results of the two approaches for one feature, each.

\begin{table}[htbp]
\centering
\begin{tabular}{lrr|rr}
\toprule
& \multicolumn{2}{c|}{Ex.~I} & \multicolumn{2}{c}{Ex.~II}\\
Bin & WoE & Pts & WoE & Pts\\
\midrule
0        & 0.396  & 6   & 0.396  & 6 \\
$\geq 1$ & -2.302 & -36 & -2.302 & -35 \\
\bottomrule
\end{tabular}
\caption{N90Late}
\label{tab:N90}
\end{table}

\begin{table}[htbp]
\centering
\begin{tabular}{lrr|rr}
\toprule
& \multicolumn{2}{c|}{Ex.~I} & \multicolumn{2}{c}{Ex.~II}\\
Bin & WoE & Pts & WoE & Pts\\
\midrule
0          & 0.532  & 8   & 0.532  & 8 \\
1          & -0.882 & -13 & -0.882 & -14 \\
$\geq 2$   & -1.895 & -28 & -1.895 & -29 \\
\bottomrule
\end{tabular}
\caption{N30--59Late}
\label{tab:N30}
\end{table}

\begin{table}[htbp]
\centering
\begin{tabular}{lrr|rr}
\toprule
& \multicolumn{2}{c|}{Ex.~I} & \multicolumn{2}{c}{Ex.~II}\\
Bin & WoE & Pts & WoE & Pts\\
\midrule
0        & 0.290  & 3   & 0.290  & 3 \\
$\geq 1$ & -2.081 & -24 & -2.081 & -25 \\
\bottomrule
\end{tabular}
\caption{N60--89Late}
\label{tab:N60}
\end{table}

As can be seen in Tables \ref{tab:N90}, \ref{tab:N30}, and \ref{tab:N60}, both approaches resulted in the same binnings for the features \textit{N90Late}, \textit{N30-59Late}, and \textit{N60-89Late}. This is easily explained by the limited number of possible bin thresholds to choose from, since these are integer-valued features.

\begin{table}[htbp]
\centering
\begin{tabular}{lrr|lrr}
\toprule
\multicolumn{3}{c|}{Ex.~I} & \multicolumn{3}{c}{Ex.~II}\\
Bin & WoE & Pts & Bin & WoE & Pts\\
\midrule
$[0,0.043]$       & 1.386 & 27 &
$[0,0.1]$         & 1.336 & 26 \\
$(0.043,0.067]$   & 1.381 & 27 &
$(0.1,0.3]$       & 0.851 & 16 \\
$(0.067,0.132]$   & 1.153 & 23 &
$(0.3,0.6]$       & -0.010 & 0 \\
$(0.132,0.184]$   & 0.932 & 18 &
$(0.6,0.9]$       & -0.774 & -15 \\
$(0.184,0.301]$   & 0.659 & 13 &
$\geq 0.9$        & -1.389 & -27 \\
$(0.301,0.396]$   & 0.234 & 5 &
& & \\
$(0.396,0.5]$     & 0.050 & 1 &
& & \\
$(0.5,0.855]$     & -0.603 & -12 &
& & \\
$(0.855,0.989]$   & -1.190 & -23 &
& & \\
$\geq 0.989$      & -1.447 & -28 &
& & \\
\bottomrule
\end{tabular}
\caption{RevUtil}
\label{tab:Rev}
\end{table}

\begin{table}[!h]
\centering
\begin{tabular}{lrr|lrr}
\toprule
\multicolumn{3}{c|}{Ex.~I} & \multicolumn{3}{c}{Ex.~II}\\
Bin & WoE & Pts & Bin & WoE & Pts\\
\midrule
Missing           & 0.211 & 3 &
Missing           & 0.211 & 2 \\
$[0,3332.5]$      & -0.355 & -5 &
$[0,3000]$        & -0.334 & -3 \\
$(3332.5,3920.5]$ & -0.246 & -3 &
$(3000,5000]$     & -0.227 & -2 \\
$(3920.5,4621.5]$ & -0.220 & -3 &
$(5000,7000]$     & 0.009 & 0 \\
$(4621.5,5563.5]$ & -0.106 & -1 &
$(7000,10000]$    & 0.231 & 2 \\
$(5563.5,6550.5]$ & 0.002 & 0 &
$\geq 10000$      & 0.464 & 5 \\
$(6550.5,7656.5]$ & 0.190 & 2 &
& & \\
$(7656.5,10284]$  & 0.286 & 4 &
& & \\
$\geq 10284$      & 0.474 & 6 &
& & \\
\bottomrule
\end{tabular}
\caption{MoInc}
\label{tab:MoInc}
\end{table}

Where more potential split points were available, namely the real-valued features \textit{RevUtil} (Table \ref{tab:Rev}) and \textit{MoInc} (Table \ref{tab:MoInc}), the bins differ significantly between the semi-automated and manual approaches.

Note that, while five of the features selected or the final model are the same across both experiments, the semi-automated approach resulted in \textit{DebtRatio} being selected as the sixth feature whereas \textit{NumRealEstate} was chosen in the manual approach.

\begin{table}[htbp]
\centering
\begin{tabular}{lrr}
\toprule
Bin & WoE & Pts\\
\midrule
$[0,0.020]$      & 0.251 & 8 \\
$(0.020,0.346]$  & 0.125 & 4 \\
$(0.346,0.423]$  & 0.048 & 1 \\
$(0.423,0.505]$  & -0.072 & -2 \\
$\geq 0.505$     & -0.162 & -5 \\
\bottomrule
\end{tabular}
\caption{DebtRatio (only Example I)}
\label{tab:debt}
\end{table}

\begin{table}[htbp]
\centering
\begin{tabular}{lrr}
\toprule
Bin & WoE & Pts\\
\midrule
0            & -0.235 & -3 \\
$\{1,2\}$    & 0.233 & 3 \\
3            & -0.067 & -1 \\
$\geq 4$     & -0.582 & -8 \\
\bottomrule
\end{tabular}
\caption{NumRealEstate (only Example II)}
\label{tab:numreal}
\end{table}

The notebooks for both approaches are available in our GitHub repository at \url{https://github.com/fraunhofer-iais/inference-framework-credit-scorecards}.

\end{document}